\documentclass[conference]{IEEEtran}

\IEEEoverridecommandlockouts
\usepackage{cite}
\ifCLASSINFOpdf
\else
\fi
\usepackage{algorithmic}
\usepackage{algorithm}

\usepackage{enumerate}
\usepackage{color}
\usepackage{graphicx}
\usepackage{epstopdf}
\usepackage{multicol}
\usepackage{mathtools}
\usepackage{amssymb,latexsym,amsmath, mathtools}
\usepackage{graphicx}
\usepackage{amssymb}
\usepackage{bbm,url}
\usepackage[keeplastbox]{flushend}
\usepackage{csquotes}
\usepackage{subcaption}

\newcommand{\code}[1]{\texttt{#1}}

\def\mb{\mathbb}
\def\mbf{\mathbf}
\def\mc{\mathcal}

\def\m{\mathbf}

\newcommand{\ip}[2]{\left\langle #1, #2 \right\rangle}
\newcommand{\br}[1]{\left[#1 \right]}
\newcommand{\pr}[1]{\left(#1 \right)}
\newcommand{\nor}[1]{\left\|#1 \right\|}
\newcommand{\trnorm}[1]{\nor{#1}_{\mathrm{tr}} }

\newcommand{\rank}{\mathop{\mathrm{rank}}\nolimits}
\newcommand{\shrink}{\mathop{\mathrm{shrink}}\nolimits}

\newcommand{\refold}{\mathop{\mathrm{refold}}\nolimits}

\newcommand{\vect}{\mathop{\mathrm{vec}}\nolimits}

\newcommand{\argmin}{\mathop{\mathrm{argmin}}}

\newcommand{\Dp}{\t D^{\pi}}

\newcommand{\const}{\mathop{\mathrm{const.}}\nolimits}
\newcommand{\stark}{{STARK}}

\renewcommand{\v}[1]{\mbf{#1}}
\renewcommand{\t}[1]{\underline{\mbf{#1}}}
\renewcommand{\m}[1]{\v{#1}}

\usepackage{setspace}

\usepackage[normalem]{ulem}
\newcommand\redout{\bgroup\markoverwith{\textcolor{red}{\rule[.5ex]{2pt}{0.4pt}}}\ULon}

\begin{document}
%
\title{STARK: Structured Dictionary Learning Through Rank-one Tensor Recovery}
\author{
\IEEEauthorblockN{Mohsen~Ghassemi, Zahra~Shakeri, Anand~D.~Sarwate,  Waheed~U.~Bajwa}
\IEEEauthorblockA{
Dept. of Electrical and Computer Engineering, Rutgers University, Piscataway, NJ 08854\\
\texttt{\{mohsen.ghassemi, zahra.shakeri, anand.sarwate, waheed.bajwa\}@rutgers.edu}
\thanks{The work of the authors was supported in part by the National Science Foundation under awards CCF-1525276 and CCF-1453073, and by the Army Research Office under award W911NF-17-1-0546.}
}
}


\maketitle

\begin{abstract}
In recent years, a class of dictionaries have been proposed for multidimensional (tensor) data representation that exploit the structure of tensor data by imposing a Kronecker structure on the dictionary underlying the data.
In this work, a novel algorithm called ``STARK'' is provided to learn Kronecker structured dictionaries that can represent tensors of any order. By establishing that the Kronecker product of any number of matrices can be rearranged to form a rank-1 tensor, we show that Kronecker structure can be enforced on the dictionary by solving a rank-1 tensor recovery problem.
Because  rank-1 tensor recovery is a challenging nonconvex problem, we resort to solving a convex relaxation of this problem.
Empirical experiments on synthetic and real data show promising results for our proposed algorithm.

\end{abstract}

\IEEEpeerreviewmaketitle

\section{Introduction}

Sparse representations of data have been widely used in a variety of information processing tasks such as data compression, feature extraction, data classification, signal
denoising and inpainting, and audio processing~\cite{kreutz2003dictionary,elad2005simultaneous,aharon2006img}. One of the powerful techniques to obtain sparse representations is dictionary learning (DL) which can be formulated as 
\begin{align}\label{standard_dictionary_problem_compact}
&\min_{\m D,\m X} ~\frac{1}{2} \sum_{i=1}^L \nor{\v y_i -\m{D} \v x_i}_2^2, \quad \text{s.t.} \quad \forall  i~\nor{\v x_i}_0\leq s.
\end{align}
We wish to find an overcomplete basis $\m D\in \mb R^{m\times p}$ with unit-norm columns and dictionary coefficient matrix $\m X =[\v x_1 ,\dots, \v x_L] \in \mb R^{p \times L}$ such that each observation $\v y_i$ 
is represented by a linear combination of no more than $s$ columns of $\m D$. Since this problem is not convex, it is typically solved by alternating minimization; $\m X$ is updated using a fixed $\m D$ and then, $\m D$ is updated using a fixed $\m X$~\cite{aharon2006img}.

In traditional DL literature, when dealing with multidimensional data, high order data $\{\t{Y}_i\}_{i=1}^L$ (tensors of order $2$ or higher) are vectorized and stacked in columns of an observation matrix $\m Y=\br{\vect(\t{Y}_1), \cdots ,\vect(\t{Y}_L)}$: the structure of the data is not considered in the dictionary underlying the data. In this case, any standard DL method can be used to find sparse representations of data. This simplistic method disregards the multidimensional structure in the data and does not capture the correlation and structure along different dimensions in each original ``data point.''

On the other hand, in structured DL methods for tensor data, the multidimentional structure in data is taken into account. There exists a class of DL algorithms for tensor data that are based on the \emph{Tucker decomposition}~\cite{tucker1963implications} of tensors. The resulting ``Kronecker structured'' DL methods (KS-DL) assume the dictionary consists of the Kronecker product~\cite{van2000ubiquitous} of smaller subdictionaries. Such algorithms represent tensor data using many fewer parameters compared to vectorized DL techniques~\cite{hawe2013separable,roemer2014tensor,dantas2017learning,zubair2013tensor}. This is due to the fact that the number of degrees of freedom in the KS-DL problem is significantly less than the traditional DL problem; this suggests that dictionary recovery is possible with smaller sample complexity using KS-DL methods \cite{ShakeriBS:16isit,shakeri2016minimax}. These works provide KS-DL algorithms to represent second order~\cite{hawe2013separable,roemer2014tensor,dantas2017learning} and 3rd-order tensors~\cite{zubair2013tensor}. In~\cite{hawe2013separable}, for example, the KS-DL objective function is minimized using a Riemannian conjugate gradient method along with a nonmonotonic line search. In other methods, the subdictionaries composing the KS dictionary are updated alternately; in~\cite{roemer2014tensor}, an approach similar to K-SVD~\cite{aharon2006img} is employed that uses higher-order SVD (HOSVD)~\cite{de2000multilinear} to alternately update coordinate dictionaries for second order tensor data and in~\cite{zubair2013tensor}, gradient descent is used to alternately update coordinate dictionaries for third order tensor data. The dictionary update stage in these algorithms involves solving a nonconvex minimization problem.
The central challenge in the theoretical analysis of such KS-DL solvers is that the dictionary update stage is nonconvex. Furthermore, these explicit models are specialized to KS problems: they do not extend to more general structures in the underlying dictionary. In contrast,
Dantas et al.~\cite{dantas2017learning} recently proposed an algorithm to learn the sum of KS dictionaries that represent second order tensor data by adding a regularizer in the objective function.

In this paper, we propose a novel algorithm called ``STARK'' to learn KS dictionaries for $N$th-order tensor data ($N\geq 2$). Our method involves adding a regularization term to the objective function of the DL problem defined in~\eqref{standard_dictionary_problem_compact}. The motivation for this term comes from the following realization: elements of any KS matrix can be rearranged to form a rank-$1$ tensor. Thus, enforcing a rank-1 constraint on such rearrangement of the dictionary results in a KS dictionary. 
To this end, we take advantage of low-rank tensor estimation literature~\cite{Gandy_2011, Romera-Paredes_multilinear, Wimalawarne_2014, huang2014provable, Liu_2013_Visual} to add a convex regularizer that imposes low-rankness on the rearrangement tensor. 
This formulation has the advantage that it can be used to learn KS dictionaries as well as the case where the underlying dictionary is better approximated by sum of KS dictionaries. 
Our method can learn dictionaries of arbitrary tensor order; in the case of second order tensor data our general formulation coincides with that of Danita's et al. \cite{dantas2017learning}.

We conduct numerical experiments to validate the performance of our algorithm. 
We use STARK for representation of third order synthetic and real tensor data and demonstrate that STARK outperforms vectorized DL technique K-SVD~\cite{aharon2006img} and KS-DL technique K-HOSVD~\cite{roemer2014tensor} for small sample sizes.

\textit{Notation Convention:} Underlined bold upper-case, bold upper-case and lower-case letters are used to denote tensors, matrices and vectors, respectively. Lower-case letters denote scalars. We denote the Kronecker product and outer product by $\otimes$ and $\circ$, respectively, while $\times_n$ denotes the mode $n$ product between a tensor and a matrix \cite{kolda_tensor}. Norms are given by subscripts, so $\|\m v\|_0$ and $\|\m v\|_2$ are the $\ell_0$ and  $\ell_2$ norms of $\m v$, while $\|\m X\|_2$, $\|\m X\|_F$, and $\|\m X \|_*$ are the spectral, Frobenius, and nuclear norms of $\m X$, respectively. A slice of a tensor is a $2$-dimensional section defined by fixing all but two of its indices. Particularly, a frontal slice of a $3$-dimensional tensor is defined by fixing the third index.


\section{Tucker-Based KS-DL}

According to the Tucker decomposition of tensors, an $N$-th order data tensor $\t Y_i\in \mb{R}^{n_1 \times n_2 \times \cdots \times n_N}$ can be decomposed in the following form:
\begin{align}
\t Y_i=& \t X_i \times_1 \m D_{1} \times_2 \m D_{2} \times_3 \cdots \times_N \m D_{N},
\end{align}
where $\t X_i\in \mb{R}^{p_1 \times p_2 \times \cdots \times p_N}$ denotes the core tensor and $\m D_{i}$'s denote transformation matrices along each mode of $\t Y_i$. The vectorized version of $\t Y_i$ 
can be written as
\begin{align}\label{tucker_kronecker}
\v y_i= \pr{\m D_{N} \otimes \m D_{N-1}\otimes \cdots \otimes \m D_{1}} \v x_i,
\end{align}
where $\v y_i=\vect(\t Y_i)$ and $\v x_i=\vect(\t X_i)$ \cite{kolda_tensor}. Here, the structure in tensor data is being exploited using the Kronecker product of transformation matrices. Stacking $L$  vectorized data points $\{\v y_i\}_{i=1}^L$ in columns of a matrix $\m Y$, we get
\begin{align}\label{Tucker_representation}
\m Y= \pr{\m D_{N} \otimes \m D_{N-1}\otimes \cdots \otimes \m D_{1}} \m X,
\end{align}
which is similar to the conventional DL problem, except that the dictionary $\m D$ is Kronecker structured.


In the next section, we present our proposed KS-DL algorithm called ``STructured dictionAry learning through RanK-1 Tensor recovery" (STARK), which implicitly enforces Kronecker structure on the dictionary being learned by means of a regularizer in the DL objective function.

We note that STARK can be used to enforce a more general structure called \textit{low-separation-rank (LSR)} structure. An LSR matrix can be written as sum of a few KS matrices: 
\begin{align}\label{eq:sum_A}
\m D=\sum_{k=1}^{K} \m D^k_{N}\otimes \m D^k_{N-1}\otimes \cdots \otimes \m D^k_{1},
\end{align}
where the factor matrices $\{\m D_i^k \in \mb{R}^{m_i \times p_i}\}_{k=1}^K$ have the same size for a fixed $i$, and $K$ is the
\textit{separation rank} of $\m D$~\cite{hero_2013_kronecker}. 



\section{Enforcing Structure via Regularization}

\begin{figure}
\centering
 \includegraphics[width=\linewidth]{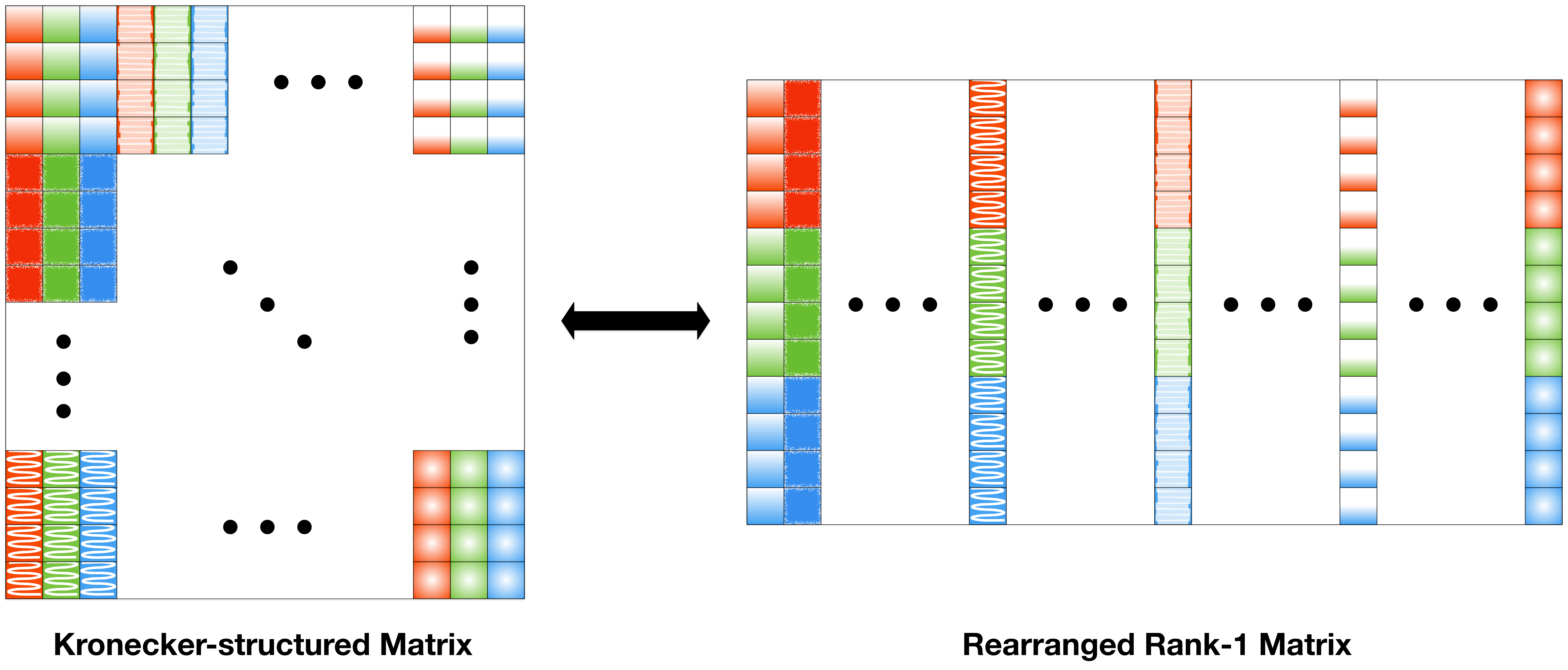}
\caption{Example of rearranging a KS matrix into a rank-1 matrix.}
\label{figure_permutation}
\end{figure}

To motivate the idea behind \stark, let us consider $\m D=\m D_1\otimes \m D_2$. It turns out that the elements of $\m D$ can be rearranged to form $\m D^{\pi}=\v d_2 \circ \v d_1$, where $\v d_i=\vect(\m D_i)$ for $i=1,2$ \cite{van2000ubiquitous}. Figure \ref{figure_permutation} illustrates this rearrangement for $\m D$.
Similarly, for $\m D=\m D_1\otimes \m D_2\otimes \m D_3$, we can write $\t{D}^{\pi}= \v d_3 \circ \v d_2\circ \v d_1$, where each frontal slice of the tensor $\t{D}^{\pi}$ is a scaled copy of $\v d_3\circ \v d_2$. Following a similar procedure, we can show that if $\m D= \sum_{k=1}^K \m D^k_{1}\otimes \m D^k_{2}\otimes \cdots \otimes \m D^k_{N}
$, then a certain ``rearrangement'' of $\m D$ is the rank-$K$ tensor
 $
\t{D}^{\pi}=\sum_{k=1}^K\v d^k_N \circ \v d^k_{n-1}\circ \cdots \circ \v d^k_1,
$ 
where $\v d_i=\vect(\m D_i)$ for $i\in [N]$. This suggests that in the structured DL problem, we can impose the LSR structure (KS when $K=1$) on the dictionary $\m D$ being learned by minimizing the rank of $\Dp$. 

Since tensor rank is a nonconvex function, in order to make this DL problem convex with respect to $\m D$, we use a commonly used convex proxy for the tensor rank function, the \textit{sum-trace-norm} \cite{Wimalawarne_2014}, which is defined as the average of the trace (nuclear) norms of the \textit{unfoldings} of the tensor:
$
\trnorm{{\t D}}=\frac{1}{N} \sum_{n=1}^{N} \nor{\m D_{(n)}}_*.
$
Using this convex relaxation for the rank function, the KS-DL problem has the following form:
\begin{align}\label{convex_structured_dictionary_problem_compact}
&\min_{\m D,\m X}~\frac{1}{2}\nor{\m Y-\m{DX}}_F^2+ \lambda \trnorm{\t{D}^{\pi}}, \;\; \text{s.t.} \;\; \forall  i~\nor{\v x_i}_0\leq s,
\end{align}
where the columns of $\m D$ have unit norm. We use alternating minimization to solve this nonconvex problem. To minimize the objective function in \eqref{convex_structured_dictionary_problem_compact} with respect to $\m X$, we can use any of the standard sparse coding methods. In simulations, we use orthogonal matching pursuit (OMP) \cite{Pati_93_omp, tropp_2007_omp}. To update the KS dictionary $\m D$, we use the alternating direction method of multipliers (ADMM) \cite{Boyd_ADMM_2011}. We describe this dictionary update step in the next section.
%

\section{Structured Dictionary Update Using ADMM}

In this section, we discuss the dictionary update step of solving problem \eqref{convex_structured_dictionary_problem_compact}, which can be stated as
\begin{align}\label{convex_structured_dictionary_update}
\min_{\m D\in \mb R^{m\times p}}~\frac{1}{2}\nor{\m Y-\m D\m X}_F^2+ \lambda \sum_{n=1}^N\nor{{\m D}^{\pi}_{(n)}}_*.
\end{align}
The main issue in solving the convex dictionary update problem \eqref{convex_structured_dictionary_update} is dealing with the interdependent nuclear norms. This makes optimization methods that use gradient information challenging. Inspired by many works in the literature on low-rank tensor estimation \cite{Romera-Paredes_multilinear, Gandy_2011,  Wimalawarne_2014, huang2014provable}, we instead suggest the following reformulation of~\eqref{convex_structured_dictionary_update}:
\begin{align}\label{dictionary_auxiliary_terms}
&\min_{\m D, \t W_1, \cdots, \t W_N}~\frac{1}{2}\nor{\m Y-\m D\m X}_F^2+ \lambda\sum_{n=1}^N\nor{(\m W_n)_{(n)}}_*\nonumber\\
&\qquad\,\text{s.t.} \quad\quad \forall n\quad \t W_n=\t{D}^{\pi}.
\end{align}
In this formulation, although the nuclear norms are associated with one another through the introduced constraint, we can decouple the minimization problem into separate subproblems. In particular, we can solve the objective function \eqref{dictionary_auxiliary_terms} using ADMM, which involves decoupling the problem into independent subproblems by forming the following augmented Lagrangian function:
\begin{align}\label{augmented_lagrange_function}
&\mc{L}_{\gamma}(\Dp, \widetilde{\t W}, \widetilde{\t A})= \frac{1}{2}\nor{\m Y-\m D\m X}_F^2+ \sum_{n=1}^N  \Big( \lambda\nor{(\m W_n)_{(n)}}_*\nonumber\\
&\qquad \quad\; \quad- \ip{\t A_n}{~\Dp-\t W_n}+\frac{\gamma}{2} \nor{~\Dp-\t W_n}_F^2 \Big),
\end{align}
where $\widetilde {\t W}=\br{\t W_1^T, \cdots,~ \t W_N^T}^T$ and $\widetilde {\t A}=\br{\t A_1^T, \cdots,~ \t A_N^T}^T$. Here, the inner product of two tensors is defined as the inner product of their vectorizations.

Finally, to find the gradient of \eqref{augmented_lagrange_function} with respect to $\Dp$, we rewrite the Lagrangian function in the following form
\begin{align}\label{augmented_lagrange_function_vector}
&\mc{L}_{\gamma}(\Dp, \widetilde{\t W}, \widetilde{\t A})= \frac{1}{2}\nor{\v y-\mc{T}(\t D^{\pi})}_2^2+ \sum_{n=1}^N  \Big( \lambda\nor{(\m W_n)_{(n)}}_*\nonumber\\
&\qquad \quad\; \quad- \ip{\t A_n}{~\Dp-\t W_n}+\frac{\gamma}{2} \nor{~\Dp-\t W_n}_F^2 \Big).
\end{align}
Here, we defined $\v y=\vect(\m Y)$ and 
$\mc{T}(\Dp)=\vect(\m D\m X)=\widetilde{\m X}^T \m \Pi \vect(\Dp)$, 
where $\widetilde{\m X}=\m X \otimes \m I_m $ and $\m \Pi$ is a permutation matrix such that $\m \Pi \vect(\Dp)=\vect(\m D)$.


In the rest of this section, we briefly discuss derivation of the permutation matrix as well as the update steps of ADMM. Due to lack of space, we leave the details to an extended version of this paper.

\subsection{The Permutation Matrix}
The permutation matrix $\m \Pi$ represents a linear transformation that maps the elements of $\vect(\m D)$ to $\vect(\Dp)$. Given index $l$ of $\vect(\m D)$ and the corresponding mapped index $l'$ of $\vect(\Dp)$, our strategy for finding the permutation matrix is to define $l'$ as a function of $l$. To this end, we first find the the corresponding row and column indices $(i,j)$ of matrix $\m D$ from the $l$th element of $\vect(\m D)$. Then, we find the index of the element of interest on the $N$th order rearranged tensor $\Dp$, and finally, we find its location $l'$ on $\vect(\Dp)$. Note that the permutation matrix is only a function of the dimensions of the factor matrices
. We leave the formal explanation of this procedure to an extended version of this paper.

\subsection{ADMM Update Rules}\label{ADMM_update}

Recall that each iteration of ADMM consists of the following steps \cite{Boyd_ADMM_2011}:
\begin{align}
&\t D^{\pi}(t+1)=\argmin\limits_{\t D^{\pi}} \mc{L}_{\gamma}  (\t D^{\pi}, \widetilde{\t W}(t), \widetilde{\t A}(t)),\label{D}\\
&\widetilde{\t W}(t+1)=\argmin\limits_{\widetilde{\t W}} \mc{L}_{\gamma}  (\t D^{\pi}(t+1), \widetilde{\t W}, \widetilde{\t A}(t)),\label{W}\\
&\widetilde{\t A}(t+1)=\widetilde{\t A}(t)- \gamma \pr{\t H\Dp(t+1)-\widetilde{\t W}(t+1)},\label{A}
\end{align}
where  $\t H$ is the vertical concatenation of $N$ instances of $\t I$, the identity operator on $\in \mb{R}^{m_1p_1 \times  \cdots \times m_N p_N}$. 
 

%
The solution to problem \eqref{D} is found by taking the gradient of $\mc{L}_{\gamma}(\cdot)$ with respect to $\t D^{\pi}$ and setting it to zero. Suppressing the iteration index $t$ for ease of notation, we have
\begin{align*}
\frac{\partial \mc{L_{\gamma}}}{\partial \Dp} = \mc{T}^*(\mc{T}(\Dp)-\v y)- \sum_{n=1}^N \t A_n +\sum_{n=1}^N \gamma \pr{\Dp-\t W_n},
\end{align*}
where $ \mc{T}^*$ denotes the adjoint of the linear operator $\mc{T}$\cite{Gandy_2011}. Setting the gradient to zero results in
\begin{align}
&\mc{T}^*(\mc{T}(\Dp)) +\gamma N ~\Dp=\mc{T}^*(\v y) + \sum_{n=1}^N  \pr{\t A_n+\gamma \t W_n}.\end{align}
Equivalently, we have,
\begin{align}
&\vect^{-1}\pr{\br{\m \Pi^T \widetilde{\m X} \widetilde{\m X}^T \m \Pi +\gamma N  \m I}\vect(\Dp)}\nonumber\\
&\qquad \qquad\quad =\vect^{-1}(\m \Pi^T \widetilde{\m X} \v y)+ \sum_{n=1}^N  \pr{\t A_n+\gamma \t W_n}.
\end{align}
Therefore, the update rule for $\Dp$ is
\begin{align}\label{Dp_update}
&\Dp(t+1)= \vect^{-1}\Big(\br{\m \Pi^{T} \widetilde{\m X} \widetilde{\m X}^T \m \Pi +\gamma N  \m I_{mp}}^{-1} \nonumber\\
&\quad \quad \cdot \Big[\m \Pi^{T} \widetilde{\m X} \v y+ \vect\Big(\sum_{n=1}^N  \pr{\t A_n(t)+\gamma\t W_n(t)}\Big)\Big]\Big).
\end{align}
To update $\widetilde{\t W}$, we can break the second step \eqref{W} into solving $N$ independent subproblems 
(suppressing the index $t$):
\begin{align}
\min_{\t W_n} ~\mc{L_W}=&\lambda \nor{(\m W_n)_{(n)}}_* - \ip{\t A_n}{~\Dp-\t W_n}\nonumber\\
&+\frac{\gamma}{2} \nor{~\Dp-\t W_n}_F^2. \nonumber
\end{align}
The objective function of this problem can be reformulated as
\begin{align}\label{W_objective}
\mc{L_W}=&\lambda \nor{(\m W_n)_{(n)}}_*+\frac{\gamma}{2} \big\|(\m W_n)_{(n)}-\big(\Dp_{(n)}-\frac{(\m A_n)_{(n)}}{\gamma}\big)\big\|_F^2 \nonumber\\
&+\const
\end{align}
The minimizer of an objective function of the form \eqref{W_objective} is
\begin{align}
\shrink\pr{(\m D^{\pi})_{(n)}-\frac{1}{\gamma} (\m A_n)_{(n)},~\frac{\lambda}{ \gamma}},
\end{align}
where $\shrink(\m M, \tau)$ is the shrinkage operator that applies soft-thresholding at level $\tau$ on the singular values of $M$ (see Theorem 2.1 in \cite{cai_2010_svt} for details).
Therefore,
\begin{align}\label{W_update}
\t W_n(t+1)=&\refold\bigg(\shrink\pr{\Big(\m D^{\pi}(t+1)}_{(n)}\nonumber\\
&\qquad\qquad \qquad-\frac{1}{\gamma}\pr{\m A_n(t)}_{(n)},~\frac{\lambda}{ \gamma}\Big)\bigg).
\end{align}
where $\refold(\cdot)$ is the inverse of the unfolding operator.
The summary of our DL method is provided in Algorithm~\ref{STARK_algorithm}.
\begin{algorithm}[b]
\caption{Structured Dictionary Learning through Rank-1 Tensor Recovery (STARK)} \label{STARK_algorithm}
\begin{algorithmic}[1]
\REQUIRE $\m Y$, $\m \Pi$,~$s>0$,~$\lambda>0$,~$\gamma>0$
\STATE \textbf{initialize:} $\m D(0)$,~$\m X(0)$, ,~$\widetilde{\t A}(0)$,  ~$\widetilde{\t W}(0)$
\STATE
\WHILE {$\nor{\m Y-\m D(\tau)\m X(\tau)}_F>\epsilon$}
\STATE \textbf{Sparse coding stage:} Use OMP to update $\m X(\tau)$.
\STATE \textbf{Dictionary update stage:}
\begin{ALC@g}
{\WHILE {$\nor{\widetilde{\t A}(t) -\widetilde{\t A}(t-1) }_F>\epsilon$}
		\STATE Update $\Dp(t)$ according to update rule \eqref{Dp_update}
		\FORALL{$i \in [N]$}
				\STATE Update $\widetilde{\t W}_n(t)$ according to update rule \eqref{W_update}			
		\ENDFOR	
		\FORALL{$n \in [N]$}
				\STATE $\t A_n(t+1)=\t A_n(t)- \gamma \pr{\Dp(t+1)-\t W_n(t+1)}$
		\ENDFOR	
\ENDWHILE}
\STATE Normalize columns of $\m D(\tau)$	
\end{ALC@g}
\ENDWHILE

\RETURN $\m D(\tau)$,~$\m X(\tau)$
\end{algorithmic}
\end{algorithm}


\section{Numerical Experiments}

We compare the performance of STARK with two methods: KSVD~\cite{aharon2006img}, as a non-structured DL method, and K-HOSVD~\cite{roemer2014tensor}, which is a structured DL method that explicitly enforces Kronecker structure on the dictionary.
We compare the performance of these methods are compared for synthetic $2$-dimensional and $3$-dimensional data as well as $3$-dimensional real-world data.

\paragraph{Synthetic Data}
For synthetic data, we randomly generate the dictionary $\m D$ and the sparse coefficient matrix $\m X$ to construct the observation matrix $\m Y=\m {DX}$.  We generate a KS dictionary as $\m D=\m D_1 \otimes \m D_2 \otimes \m D_3$ (and $\m D=\m D_1 \otimes \m D_2$ for $2$-dimensional data) with unit-norm columns according to the following procedure. The elements of the subdictionaries $\m D_1$ through $\m D_3$ are chosen i.i.d from a Gaussian distribution $\mc{N}(0,1)$, and then the columns of the subdictionaries are normalized. For generating $\m X$, we select the locations of the $s$ nonzero elements of each column uniformly at random. The values of those elements are sampled i.i.d. from $\mc N(0,1)$. In the learning process, the dictionary $\m D$ is initialized using random columns of the observation matrix $\m Y$. 
The experiments were run for 20 Monte Carlo iterations and for various training sample sizes and the resulting dictionaries were tested on a set of $10000$ test samples.   For 2nd-order tensor data we selected $m_1=4$,  $p_1=12$, $m_2=6$, $p_2=8$, $s=5$,
and for 3rd-order tensor data we selected $m_1=2$,  $p_1=4$, $m_2=5$, $p_2=10$, $m_3=5$, $p_3=5$, and $s=10$.
%
%

\begin{figure}
\centering
 \includegraphics[width=0.48\textwidth]{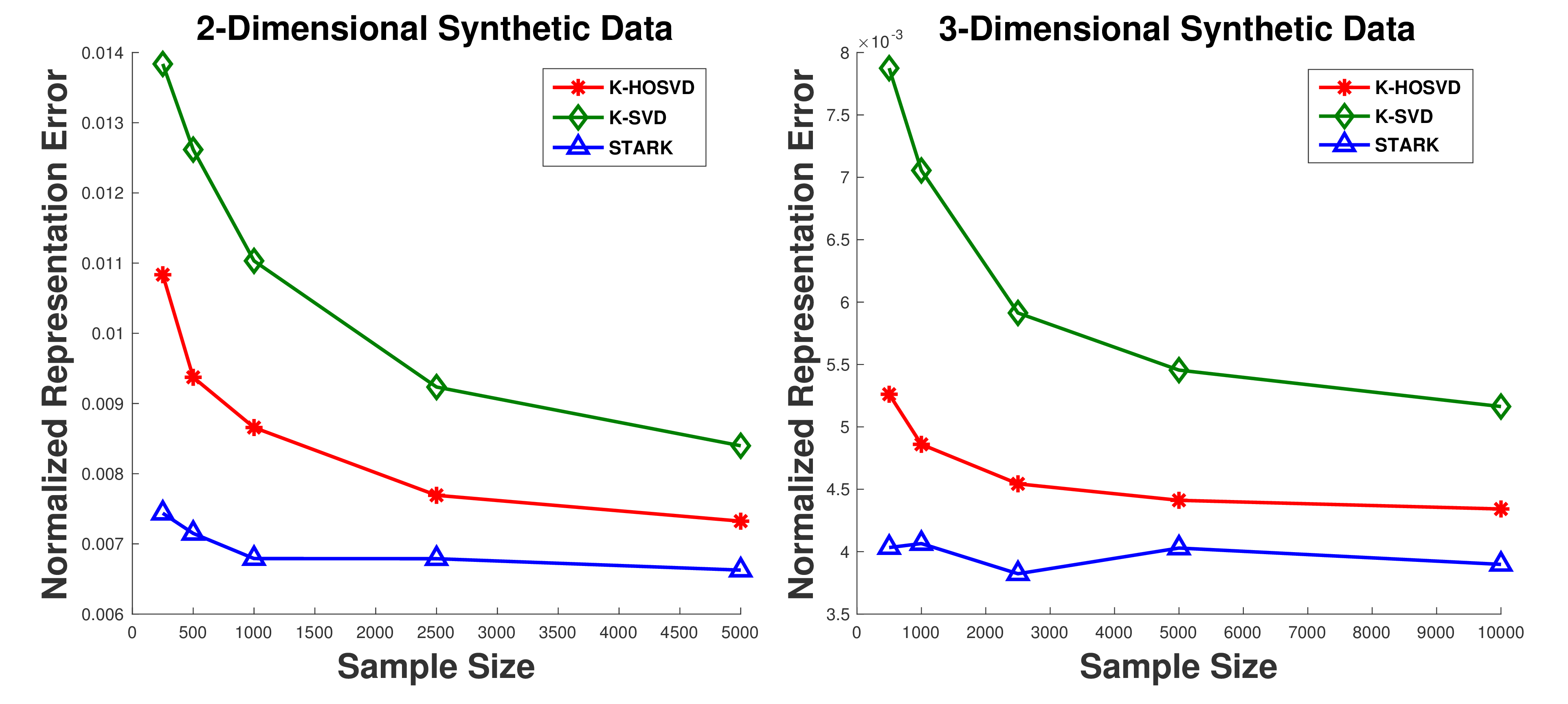}
\caption{Normalized Representation Error on Synthetic Data.} 
\label{Synthetic_plot}
\end{figure}

The results of our experiments on synthetic data are shown in Figure \ref{Synthetic_plot}. We compared our method to K-SVD and K-HOSVD. For K-HOSVD, we used algorithm provided by the authors, which actually enforces a Khatri-Rao structure on the dictionary rather than KS. STARK outperforms both K-SVD and K-HOSVD for all training sample sizes, especially when the number of training samples is small. The improvement over K-SVD can be attributed to the lower sample complexity of structured DL models. We conjecture that one reason for the improvement over K-HOSVD is that the dictionary update in STARK is a convex problem and thus the algorithm is less prone to getting stuck in a poor local minimum.


\paragraph{Real Data}
For these experiments, we compare the denoising performance of the three methods on two RGB images, \code{Peppers} and \code{Lena}, which are $512\times 512\times 2$ and $512\times 512\times 3$ tensors, respectively. We corrupt the images using additive white Gaussian noise with $\sigma=50$.
To construct the training data set, we extract overlapping $6\times 6$ patches from each image and treat each patch as a data point. Then we compare the denoising performances of the methods based on the resulting peak signal to noise ratio (PSNR) of the reconstructed images~\cite{hore2010psnr}.
%

\begin{figure}
\centering
 \includegraphics[width=0.48\textwidth]{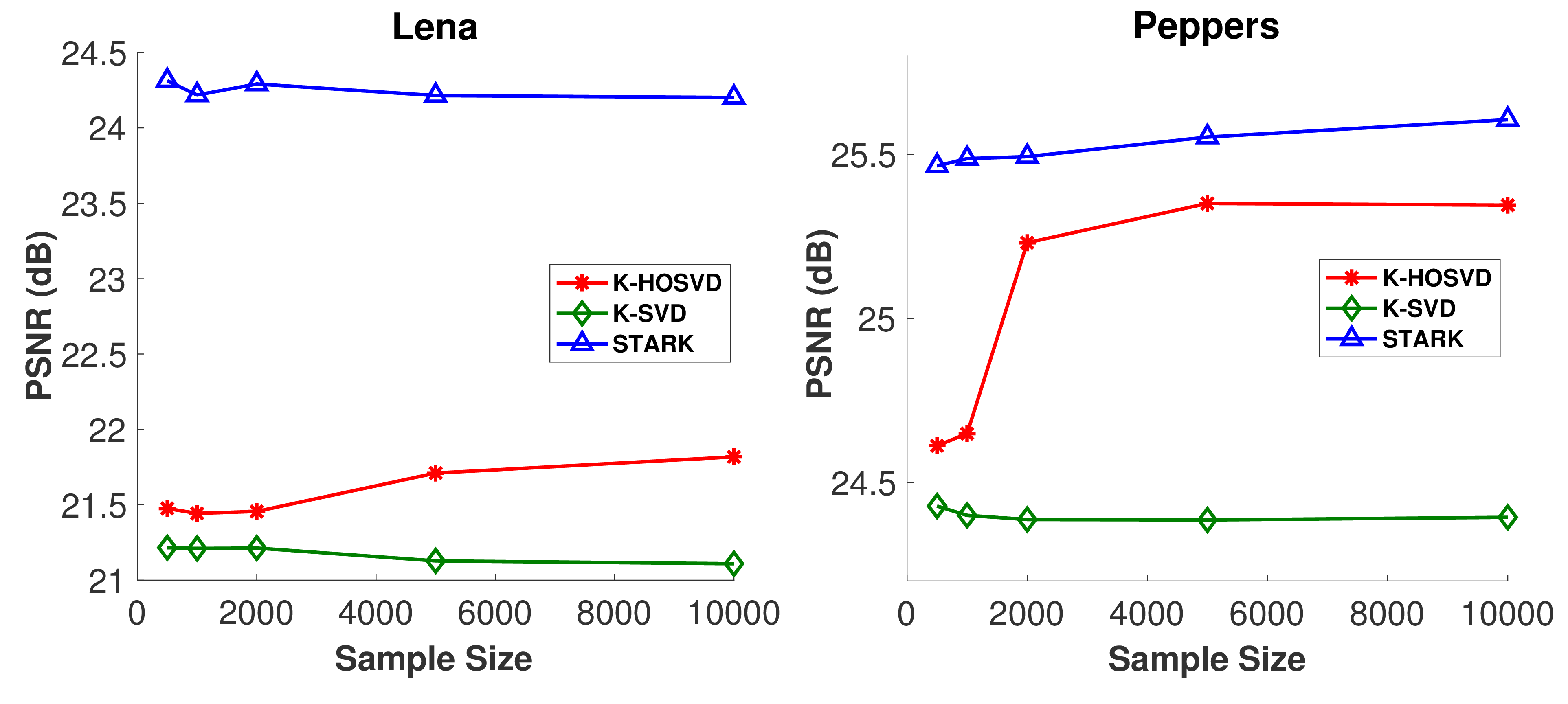}
\caption{Denoising Performance on Real Data (PSNR).}
\label{real_plot}
\end{figure}

Figure \ref{real_plot} shows the results averaged over 10 Monte Carlo iterations. We can see the denoising performance of STARK is superior to both K-SVD and K-HOSVD for all training sample sizes. This is in part due to the fact that for real-world data, the underlying dictionaries may not be KS. STARK allows $\Dp$ to have rank higher than $1$, meaning the algorithm can use a larger number of parameters ($K$ times as many when $\rank(\Dp)=K$) to approximate the true dictionary.

\section{Conclusion}
In this paper we showed that the Kronecker product of $N$ matrices can be rearranged to form an $N$th order rank-1 tensor. Based on this, we proposed a novel structured dictionary learning method for multidimensional data that enforces LSR structure in the dictionary through imposing a low-rank structure on the rearranged tensor. In particular, Kronecker structure can be enforced by imposing a rank-1 constraint on the rearranged tensor.  Since the low-rank tensor recovery problem is a nonconvex problem, we resort to solving its convex relaxation, namely, minimizing the sum-trace-norm of the rearranged tensor, which is a convex proxy for tensor rank.
We used ADMM for solving the dictionary update stage of this structured DL problem. Our experiments on both synthetic and real data showed that when the sample size is small, our method considerably outperforms both K-SVD, which returns unstructured dictionaries, and K-HOSVD, a KS-DL method that directly finds the subdictionaries.
\bibliographystyle{IEEEtran}
\bibliography{IEEEabrv,ref}

\begin{thebibliography}{10}
\providecommand{\url}[1]{#1}
\csname url@samestyle\endcsname
\providecommand{\newblock}{\relax}
\providecommand{\bibinfo}[2]{#2}
\providecommand{\BIBentrySTDinterwordspacing}{\spaceskip=0pt\relax}
\providecommand{\BIBentryALTinterwordstretchfactor}{4}
\providecommand{\BIBentryALTinterwordspacing}{\spaceskip=\fontdimen2\font plus
\BIBentryALTinterwordstretchfactor\fontdimen3\font minus
  \fontdimen4\font\relax}
\providecommand{\BIBforeignlanguage}[2]{{%
\expandafter\ifx\csname l@#1\endcsname\relax
\typeout{** WARNING: IEEEtran.bst: No hyphenation pattern has been}%
\typeout{** loaded for the language `#1'. Using the pattern for}%
\typeout{** the default language instead.}%
\else
\language=\csname l@#1\endcsname
\fi
#2}}
\providecommand{\BIBdecl}{\relax}
\BIBdecl

\bibitem{kreutz2003dictionary}
\BIBentryALTinterwordspacing
K.~Kreutz-Delgado, J.~F. Murray, B.~D. Rao, K.~Engan, T.-W. Lee, and T.~J.
  Sejnowski, ``Dictionary learning algorithms for sparse representation,''
  \emph{Neural computation}, vol.~15, no.~2, pp. 349--396, February 2003.
  [Online]. Available: \url{https://doi.org/10.1162/089976603762552951}
\BIBentrySTDinterwordspacing

\bibitem{elad2005simultaneous}
\BIBentryALTinterwordspacing
M.~Elad, J.-L. Starck, P.~Querre, and D.~L. Donoho, ``Simultaneous cartoon and
  texture image inpainting using morphological component analysis ({MCA}),''
  \emph{Appl. and Computational Harmonic Anal.}, vol.~19, no.~3, pp. 340--358,
  November 2005. [Online]. Available:
  \url{https://doi.org/10.1016/j.acha.2005.03.005}
\BIBentrySTDinterwordspacing

\bibitem{aharon2006img}
\BIBentryALTinterwordspacing
M.~Aharon, M.~Elad, and A.~Bruckstein, ``{$K$-SVD}: An algorithm for designing
  overcomplete dictionaries for sparse representation,'' \emph{IEEE Trans.
  Signal Process.}, vol.~54, no.~11, pp. 4311--4322, November 2006. [Online].
  Available: \url{https://doi.org/10.1109/TSP.2006.881199}
\BIBentrySTDinterwordspacing

\bibitem{tucker1963implications}
L.~R. Tucker, ``Implications of factor analysis of three-way matrices for
  measurement of change,'' \emph{Problems in Measuring Change}, pp. 122--137,
  1963.

\bibitem{van2000ubiquitous}
\BIBentryALTinterwordspacing
C.~F. Van~Loan, ``The ubiquitous {Kronecker} product,'' \emph{J. Computational
  and Appl. Math.}, vol. 123, no.~1, pp. 85--100, November 2000. [Online].
  Available: \url{https://doi.org/10.1016/S0377-0427(00)00393-9}
\BIBentrySTDinterwordspacing

\bibitem{hawe2013separable}
\BIBentryALTinterwordspacing
S.~Hawe, M.~Seibert, and M.~Kleinsteuber, ``Separable dictionary learning,'' in
  \emph{Proc. IEEE Conf. Comput. Vision and Pattern Recognition (CVPR)}, June
  2013, pp. 438--445. [Online]. Available:
  \url{https://doi.org/10.1109/CVPR.2013.63}
\BIBentrySTDinterwordspacing

\bibitem{roemer2014tensor}
\BIBentryALTinterwordspacing
F.~Roemer, G.~Del~Galdo, and M.~Haardt, ``Tensor-based algorithms for learning
  multidimensional separable dictionaries,'' in \emph{Proc. IEEE Int. Conf.
  Acoustics, Speech and Signal Process. (ICASSP)}, May 2014, pp. 3963--3967.
  [Online]. Available: \url{https://doi.org/10.1109/ICASSP.2014.6854345}
\BIBentrySTDinterwordspacing

\bibitem{dantas2017learning}
\BIBentryALTinterwordspacing
C.~F. Dantas, M.~N. da~Costa, and R.~da~Rocha~Lopes, ``Learning dictionaries as
  a sum of {K}ronecker products,'' \emph{IEEE Signal Process. Lett.}, vol.~24,
  no.~5, pp. 559--563, March 2017. [Online]. Available:
  \url{https://doi.org/10.1109/LSP.2017.2681159}
\BIBentrySTDinterwordspacing

\bibitem{zubair2013tensor}
\BIBentryALTinterwordspacing
S.~Zubair and W.~Wang, ``Tensor dictionary learning with sparse {T}ucker
  decomposition,'' in \emph{Proc. IEEE 18th Int. Conf. Digital Signal Process.
  (DSP)}, July 2013, pp. 1--6. [Online]. Available:
  \url{https://doi.org/10.1109/ICDSP.2013.6622725}
\BIBentrySTDinterwordspacing

\bibitem{ShakeriBS:16isit}
\BIBentryALTinterwordspacing
Z.~Shakeri, W.~U. Bajwa, and A.~D. Sarwate, ``Minimax lower bounds for
  {Kronecker}-structured dictionary learning,'' in \emph{Proc. 2016 IEEE Int.
  Symp. Inf. Theory}, July 2016, pp. 1148--1152. [Online]. Available:
  \url{https://doi.org/10.1109/ISIT.2016.7541479}
\BIBentrySTDinterwordspacing

\bibitem{shakeri2016minimax}
\BIBentryALTinterwordspacing
------, ``Minimax lower bounds on dictionary learning for tensor data,''
  \emph{arXiv preprint arXiv:1608.02792}, August 2016. [Online]. Available:
  \url{https://arxiv.org/abs/1608.02792}
\BIBentrySTDinterwordspacing

\bibitem{de2000multilinear}
\BIBentryALTinterwordspacing
L.~De~Lathauwer, B.~De~Moor, and J.~Vandewalle, ``A multilinear singular value
  decomposition,'' \emph{SIAM J. Matrix Analy. and Applicat.}, vol.~21, no.~4,
  pp. 1253--1278, 2000. [Online]. Available:
  \url{https://doi.org/10.1137/S0895479896305696}
\BIBentrySTDinterwordspacing

\bibitem{Gandy_2011}
\BIBentryALTinterwordspacing
S.~Gandy, B.~Recht, and I.~Yamada, ``Tensor completion and low-n-rank tensor
  recovery via convex optimization,'' \emph{Inverse Problems}, vol.~27, no.~2,
  p. 025010, January 2011. [Online]. Available:
  \url{https://doi.org/10.1088/0266-5611/27/2/025010}
\BIBentrySTDinterwordspacing

\bibitem{Romera-Paredes_multilinear}
\BIBentryALTinterwordspacing
B.~Romera-Paredes, H.~Aung, N.~Bianchi-Berthouze, and M.~Pontil, ``Multilinear
  multitask learning,'' in \emph{Proc. 30th Int. Conf. Mach. Learn. (ICML)},
  vol.~28, no.~3, Atlanta, Georgia, USA, June 2013, pp. 1444--1452. [Online].
  Available: \url{http://proceedings.mlr.press/v28/romera-paredes13.html}
\BIBentrySTDinterwordspacing

\bibitem{Wimalawarne_2014}
\BIBentryALTinterwordspacing
K.~Wimalawarne, M.~Sugiyama, and R.~Tomioka, ``Multitask learning meets tensor
  factorization: Task imputation via convex optimization,'' in \emph{Proc.
  Advances in Neural Inform. Process. Syst. (NIPS)}, 2014, pp. 2825--2833.
  [Online]. Available: \url{http://dl.acm.org/citation.cfm?id=2969033.2969142}
\BIBentrySTDinterwordspacing

\bibitem{huang2014provable}
\BIBentryALTinterwordspacing
B.~Huang, C.~Mu, D.~Goldfarb, and J.~Wright, ``Provable low-rank tensor
  recovery,'' \emph{Optimization-Online}, vol. 4252, p.~2, February 2014.
  [Online]. Available:
  \url{http://www.optimization-online.org/DB_FILE/2014/02/4252.pdf}
\BIBentrySTDinterwordspacing

\bibitem{Liu_2013_Visual}
\BIBentryALTinterwordspacing
J.~Liu, P.~Musialski, P.~Wonka, and J.~Ye, ``Tensor completion for estimating
  missing values in visual data,'' \emph{IEEE Trans. Pattern Anal. Mach.
  Intell.}, vol.~35, no.~1, pp. 208--220, January 2013. [Online]. Available:
  \url{http://doi.org/10.1109/TPAMI.2012.39}
\BIBentrySTDinterwordspacing

\bibitem{kolda_tensor}
\BIBentryALTinterwordspacing
T.~G. Kolda and B.~W. Bader, ``Tensor decompositions and applications,''
  \emph{SIAM Review}, vol.~51, no.~3, pp. 455--500, August 2009. [Online].
  Available: \url{https://doi.org/10.1137/07070111X}
\BIBentrySTDinterwordspacing

\bibitem{hero_2013_kronecker}
\BIBentryALTinterwordspacing
T.~Tsiligkaridis and A.~O. Hero, ``Covariance estimation in high dimensions via
  {K}ronecker product expansions,'' \emph{IEEE Trans. Signal Process.},
  vol.~61, no.~21, pp. 5347--5360, November 2013. [Online]. Available:
  \url{https://doi.org/10.1109/TSP.2013.2279355}
\BIBentrySTDinterwordspacing

\bibitem{Pati_93_omp}
\BIBentryALTinterwordspacing
Y.~C. Pati, R.~Rezaiifar, and P.~S. Krishnaprasad, ``Orthogonal matching
  pursuit: recursive function approximation with applications to wavelet
  decomposition,'' in \emph{Proc. 27th Asilomar Conf. Signals, Syst. and
  Comput.}, November 1993, pp. 40--44 vol.1. [Online]. Available:
  \url{https://doi.org/10.1109/ACSSC.1993.342465}
\BIBentrySTDinterwordspacing

\bibitem{tropp_2007_omp}
\BIBentryALTinterwordspacing
J.~A. Tropp and A.~C. Gilbert, ``Signal recovery from random measurements via
  orthogonal matching pursuit,'' \emph{IEEE Trans. Inf. Theory}, vol.~53,
  no.~12, pp. 4655--4666, December 2007. [Online]. Available:
  \url{https://doi.org/10.1109/TIT.2007.909108}
\BIBentrySTDinterwordspacing

\bibitem{Boyd_ADMM_2011}
\BIBentryALTinterwordspacing
S.~Boyd, N.~Parikh, E.~Chu, B.~Peleato, and J.~Eckstein, ``Distributed
  optimization and statistical learning via the alternating direction method of
  multipliers,'' \emph{Found. Trends Mach. Learn.}, vol.~3, no.~1, pp. 1--122,
  January 2011. [Online]. Available: \url{https://doi.org/10.1561/2200000016}
\BIBentrySTDinterwordspacing

\bibitem{cai_2010_svt}
\BIBentryALTinterwordspacing
J.-F. Cai, E.~J. Cand\`{e}s, and Z.~Shen, ``A singular value thresholding
  algorithm for matrix completion,'' \emph{SIAM J. Optimization}, vol.~20,
  no.~4, pp. 1956--1982, March 2010. [Online]. Available:
  \url{https://doi.org/10.1137/080738970}
\BIBentrySTDinterwordspacing

\bibitem{hore2010psnr}
\BIBentryALTinterwordspacing
A.~Hore and D.~Ziou, ``Image quality metrics: {PSNR} vs. {SSIM},'' in
  \emph{Proc. IEEE int. conf. Pattern recognition (ICPR)}, August 2010, pp.
  2366--2369. [Online]. Available: \url{https://doi.org/10.1109/ICPR.2010.579}
\BIBentrySTDinterwordspacing

\end{thebibliography}

\end{document}